\documentclass[runningheads]{llncs}
\pdfoutput=1

\usepackage{multirow}
\usepackage{color}
\usepackage{times}
\usepackage{latexsym} 
\usepackage{amssymb} 
\usepackage{url}
\usepackage{xspace}

\newcommand{\ruleimpl}{\ensuremath{\mathtt{<--}}}
\newcommand{\fail}{\ensuremath{\mathtt{fail}}}
\newcommand{\DATABASE}{\ensuremath{\mathtt{DATABASE}}}
\newcommand{\SPECIFICATION}{\ensuremath{\mathtt{SPECIFICATION}}}
\newcommand{\figrule}{\begin{center}\hrule\end{center}}

\title{Translating NP-SPEC into ASP%
\thanks{This work was supported by M.I.U.R.\ within the PRIN project LoDeN.}
}
\author{Mario Alviano and Wolfgang Faber}

\institute{Department of Mathematics\\
           University of Calabria\\
           87030 Rende (CS), Italy \\
           \email{mario@alviano.com, wf@wfaber.com}
}
\titlerunning{Translating NP-SPEC into ASP}
\authorrunning{M.~Alviano and W.~Faber}

\begin{document}

\setcounter{page}{3}

\maketitle

\begin{abstract}
NP-SPEC is a language for specifying problems in NP in a declarative way.
Despite the fact that the semantics of the language was given by referring to
Datalog with circumscription, which is very close to ASP, so far the only
existing implementations are by means of $ECL^iPS^e$ Prolog and via Boolean
satisfiability solvers. In this paper, we present translations from NP-SPEC
into various forms of ASP and analyze them. We also argue that it might be
useful to incorporate certain language constructs of NP-SPEC into mainstream
ASP.
\end{abstract}

\section{Introduction}\label{sec:intro}
NP-SPEC is a language that was proposed in
\cite{cado-etal-1999-padl,cado-etal-2000} in order to specify problems
in the complexity class NP in a simple, clear, and declarative
way. The language is based on Datalog with circumscription, in which some predicates are
circumscribed, while others are not and are thus ``left open''. Some
practical features are added to this basic language, often by means of
reductions.

The original software system supporting NP-SPEC was described in
\cite{cado-etal-2000} and was written in the $ECL^iPS^e$ Constraint
Programming System, based on Prolog. A second software system,
SPEC2SAT\footnote{\url{http://www.dis.uniroma1.it/cadoli/research/projects/NP-SPEC/code/SPEC2SAT/}}, was proposed in \cite{cado-etal-2005-rcra}, which rewrites
NP-SPEC into propositional formulas for testing
satisfiability. The
system has also been tested quite extensively in
\cite{cado-scha-2005-aij}, also for several problems taken from
CSPLIB, with promising results.

Interestingly, to our knowledge so far no attempt has been made to
translate NP-SPEC into Answer Set Programming (ASP), which is very similar in spirit to Datalog
with circumscription, and thus a good candidate as a transformation
target. Moreover, several efficient ASP software systems are available,
which should guarantee good performance. A crucial advantage of ASP
versus propositional satisfiability is the fact that NP-SPEC
problem descriptions are in general not propositional, and therefore a
reduction from NP-SPEC to SAT has to include an implicit instantiation
(or grounding) step. Also ASP allows for variables, and ASP systems
indeed provide optimized grounding procedures, which include many
advanced techniques from database theory (such as indexing,
join-ordering, etc). This takes the burden of instantiating in a smart
way from the NP-SPEC translation when using ASP systems.

In this paper we provide a translation from NP-SPEC into various
variants of ASP. We discuss properties and limitations of the
translation and also provide a prototype implementation, for 
which we provide a preliminary experimental analysis, which shows that our approach is advantageous, in particular that it pays off if grounding tasks are delegated to existing systems.
The rest of
the paper is structured as follows: in section~\ref{sec:prelim} we
review the language NP-SPEC and give a very brief account of ASP. In
section~\ref{sec:npspectoasp} we provide the main ingredients for
translations from NP-SPEC to ASP, and discuss properties and
limitations. In section~\ref{sec:experiments} we report on preliminary
experimental results. Finally, in section~\ref{sec:concl} we draw our
conclusions.

\section{Preliminaries: NP-SPEC and ASP}\label{sec:prelim}

We first provide a brief definition of NP-SPEC programs. For details,
we refer to \cite{cado-etal-2000}. We also note that a few minor
details in the input language of SPEC2SAT (in which the publicly
available examples are written) are different to what is described in
\cite{cado-etal-2000}. We will usually stick to the syntax of
SPEC2SAT.

An NP-SPEC program consists of two main sections\footnote{SPEC2SAT
  also has a third, apparently undocumented section called
  $\mathtt{SEARCH}$, which seems to define only output features and
  which we will not describe here.}: one section called \DATABASE{} and one
called \SPECIFICATION{}, each of which is preceded by the
respective keyword.

\subsection{\DATABASE}

 The database section defines extensional
predicates or relations and (interpreted) constants. Extensional
predicates are defined by writing
\[
\mathtt{p = \{ t_1, \ldots, t_n\};}
\]
where $\mathtt{p}$ is a predicate symbol and each $\mathtt{t_i}$ is a tuple with
matching arity. For unary predicates, each tuple is simply an integer
or a constant symbol; for arity greater than 1, it is a
comma-separated sequence of integers or constant symbols enclosed in
round brackets. Unary extensions that are ranges of integers can also be abbreviated to $\mathtt{n..m}$, where $\mathtt{n}$ and $\mathtt{m}$ are integers or interpreted constants. Constant definitions are written as
\[
\mathtt{c = i;}
\]
where $\mathtt{i}$ is an integer.

\begin{example}\label{ex:graph}
The following defines the predicate $\mathtt{edge}$ representing a graph with six nodes and nine edges, and a constant $\mathtt{n}$ representing the number of nodes.
\[
\begin{array}{l}
\DATABASE\\
\quad    \mathtt{n = 6;}\\
\quad    \mathtt{edge = \{(1,2),(3,1),(2,3),(6,2),(5,6),(4,5),(3,5),(1,4),(4,1)\};}
\end{array}
\]
\end{example}

\subsection{\SPECIFICATION}

The \SPECIFICATION{} section consists of two parts: a search space declaration and a stratified Datalog program. The search space declaration serves as a domain definition for ``guessed'' predicates and must be one or more of the \emph{metafacts} $\mathtt{Subset(d,p)}$, $\mathtt{Permutation(d,p)}$, $\mathtt{Partition(d,p,n)}$, and $\mathtt{IntFunc(d,p,n..m)}$, which we will describe below.

\subsubsection{$\mathtt{Subset(d,p)}$.}
This is the basic construct to which all following search space
declaration constructs are reduced in the semantic definition in
\cite{cado-etal-2000}. Here, $\mathtt{d}$ is a \emph{domain definition}, which
is either an extensional predicate, a range $\mathtt{n..m}$, or a
Cartesian product ($\mathtt{><}$), union ($\mathtt{+}$), intersection ($\mathtt{*}$), or
difference ($\mathtt{-}$) of two domains. Symbol $\mathtt{p}$ is a predicate identifier and the intended meaning is
that the extension of $\mathtt{p}$ can be any subset of the domain definition's
extension, thus giving rise to nondeterminism or a ``guess''.

\begin{example}\label{ex:subgraph}
Together with the code of Example~\ref{ex:graph}, the following specification will represent all subgraphs (including the original graph) as extensions of predicate $\mathtt{subgraph}$.
\[
\begin{array}{l}
\SPECIFICATION\\
\quad    \mathtt{Subset(edge,subgraph).}
\end{array}
\]
\end{example}

\subsubsection{$\mathtt{Permutation(d,p)}$.}
Concerning this construct, $\mathtt{d}$ is again a domain definition, and $\mathtt{p}$ will
have an extension in which each tuple of $\mathtt{d}$ is present and an
additional argument associates a unique integer between 1 and the
cardinality of the extension of $\mathtt{d}$ (say, $\mathtt{c}$) to each tuple, thereby defining a
permutation. The extensions of $\mathtt{p}$ thus define a bijective functions from tuples of the extension of $\mathtt{d}$ to $\mathtt{\{1..c\}}$.

\begin{example}\label{ex:edgeorder}
Together with the code of Example~\ref{ex:graph}, the following specification will represent all enumerations of edges.
\[
\begin{array}{l}
\SPECIFICATION\\
\quad    \mathtt{Permutation(edge,edgeorder).}
\end{array}
\]
One extension of $\mathtt{edgeorder}$ that reflects the ordering of the edges as written in Example~\ref{ex:graph} is 
\[
\begin{array}{ccc}
 \mathtt{edgeorder(1,2,1)}, & \mathtt{edgeorder(3,1,2)}, & \mathtt{edgeorder(2,3,3)},\\
 \mathtt{edgeorder(6,2,4)}, & \mathtt{edgeorder(5,6,5)}, & \mathtt{edgeorder(4,5,6)},\\
 \mathtt{edgeorder(3,5,7)}, & \mathtt{edgeorder(1,4,8)}, & \mathtt{edgeorder(4,1,9)}.
\end{array}
\]
\end{example}

\subsubsection{$\mathtt{Partition(d,p,n)}$.}
Also in this case $\mathtt{p}$ will have one argument more than $\mathtt{d}$. In this case, extensions of $\mathtt{p}$ will define functions from tuples of the extension of $\mathtt{d}$ to $\mathtt{\{1..n\}}$, thereby defining $\mathtt{n}$ (possibly empty) partitions.

\begin{example}\label{ex:partition}
Together with the code of Example~\ref{ex:graph}, the following specification will represent all possible pairs of graphs that partition the input graph.
\[
\begin{array}{l}
\SPECIFICATION\\
\quad    \mathtt{Partition(edge,partition,2).}
\end{array}
\]
One extension of $\mathtt{partition}$ that has the first four edges in the first partition (i.e., partition 0) and the last five edges in the second partition (i.e., partition 1) would be 
\[
\begin{array}{ccc}
 \mathtt{partition(1,2,0)}, & \mathtt{partition(3,1,0)}, & \mathtt{partition(2,3,0)},\\
 \mathtt{partition(6,2,0)}, & \mathtt{partition(5,6,1)}, & \mathtt{partition(4,5,1)},\\
 \mathtt{partition(3,5,1)}, & \mathtt{partition(1,4,1)}, & \mathtt{partition(4,1,1)}.
\end{array}
\]
\end{example}

\subsubsection{$\mathtt{IntFunc(d,p,n..m)}$.}

Again, $\mathtt{p}$ will have one argument more than $\mathtt{d}$. Here, extensions of $\mathtt{p}$ will define functions from tuples of the extension of $\mathtt{d}$ to $\mathtt{\{n..m\}}$.

\begin{example}\label{ex:partitionintfunc}
The following specification is equivalent to the one in Example~\ref{ex:partition}:
\[
\begin{array}{l}
\SPECIFICATION\\
\quad    \mathtt{IntFunc(edge,partition,0..1).}
\end{array}
\]
\end{example}

\subsubsection{Stratified Datalog Program.}

The stratified Datalog program is written using $<--$ as the rule implication symbol. It may contain built-in
predicates ($==$, $<$, $>$, $>=$, $<=$, $!=$), arithmetic expressions,
and stratified aggregates ($\mathtt{COUNT}, \mathtt{SUM},
\mathtt{MIN}, \mathtt{MAX}$). It may also contain integrity
constraints, in which case rule heads contain the special symbol
$\fail$. Rule implication is denoted by $\ruleimpl$, the aggregates
are written as for example $\mathtt{SUM(p(*,\_,Y),Z:n..m)}$ where:
$\mathtt{*}$ specifies the argument to be aggregated over; variables
that are not shared with other rule literals are local (as a special
case the anonymous variable $\mathtt{\_}$) and represent the arguments
that are not fixed; variables that are shared with other rule literals
are considered fixed in the aggregation; and variable $\mathtt{Z}$ will
contain the valuation of the aggregate, which will must be in the range
$\mathtt{n..m}$. Comments may be written in
\texttt{C++} style (using $\mathtt{/* */}$ or $\mathtt{//}$).

\begin{example}
As an example, consider the well-known Hamiltonian Cycle problem. The
NP-SPEC distribution contains an example program for an example graph:
\[
\mathtt{
\begin{array}{l}
\DATABASE\\
\quad    \mathtt{n = 6;}\ //\mathtt{no.\ of\ nodes}\\
\quad    \mathtt{edge = \{(1,2),(3,1),(2,3),(6,2),(5,6),(4,5),(3,5),(1,4),(4,1)\};}\\
\SPECIFICATION\\
\quad    \mathtt{Permutation(\{1..n\},path).}\\
\quad    \mathtt{fail\ <--\ path(X,P),\ path(Y,P+1),\ NOT\ edge(X,Y).}\\
\quad    \mathtt{fail\ <--\ path(X,n),\ path(Y,1),\ NOT\ edge(X,Y).} 
\end{array}
}
\]
The \DATABASE{} section contains an encoding of the example graph by
means of the binary predicate $\mathtt{edge}$ and defines a constant
$\mathtt{n}$ for representing the number of nodes of that
graph. Implicitly it is assumed that the nodes are labeled by integers
from $1$ to $\mathtt{n}$. The \SPECIFICATION{} section then first
guesses a permutation of the nodes and then verifies the Hamiltonian
Cycle condition by means of integrity constraints, one exploiting the
linear order of the permutation identifiers, and another one to close
the cycle from the last permutation identifier to the first one.
\end{example}

The semantics of NP-SPEC programs is provided by means of Datalog with
Circumscription, in which some predicates are minimized. That means
that among all models only those which are minimal with respect to the
minimized predicates are accepted. Moreover, among these only those
which make the special symbol $\fail$ false are considered and
referred to as answers. All metafacts are reduced to the basic
metafact $\mathtt{Subset}$ that effectively states that the predicate
defined by the metafact is not minimized. For further details of the
semantics, we refer to \cite{cado-etal-2000}.

Concerning ASP, we only give a very brief overview, details may be found in works such as \cite{bara-2002,leon-etal-2006-tocl,gebs-etal-2011-aicom}. An ASP program consists of rules
\[
\mathtt{L_1 \vee \cdots \vee L_k\ :-\ Body}
\]
where the $\mathtt{L_i}$ are literals containing variables and
constants\footnote{Many modern ASP systems also allow for function
  symbols, but they are not needed here.} (possibly containing strong
negation) and $\mathtt{Body}$, which is a conjunction of literals,
that may also contain built-ins, aggregates and default negation. Rules without heads act like integrity constraints. The semantics is based
on the Gelfond-Lifschitz reduct \cite{gelf-lifs-91} and also
guarantees minimality of the answer sets.

Practical ASP systems differ in several details, for instance several
do not support disjunction in rule heads, built-in predicates and
arithmetic expressions may differ and also aggregates are sometimes
written in slightly different ways. In this paper, we will use the
syntax of gringo (\url{http://potassco.sourceforge.net/}) and DLV
(\url{http://www.dlvsystem.com}). Both systems assume that the input
programs are safe, that is, each variable in a rule must also occur in
a positive body atom. While gringo can also parse disjunctive
programs, clasp, the solver it is often used with, can only deal with
nondisjunctive programs.

\begin{example}
As an example, consider the Hamiltonian Cycle problem and instance from above. An ASP encoding similar to the NP-SPEC program seen earlier would be:
\[
\mathtt{
\begin{array}{l}
\mathtt{\#const\ n=6}\\
\mathtt{edge(1,2).\ edge(3,1).\ edge(2,3).\ edge(6,2).\ edge(5,6).}\\
\mathtt{edge(4,5).\ edge(3,5).\ edge(1,4).\ edge(4,1).}\\
\mathtt{d(1..n).}\\
\mathtt{path(X,1)\!\vee\!path(X,2)\!\vee\!path(X,3)\!\vee\!path(X,4)\!\vee\!path(X,5)\!\vee\!path(X,6) :-\ d(X).}\\
\mathtt{:-\ path(X,A),\ path(Y,A),\ X\ !=\ Y.}\\
\mathtt{:-\ path(X,P),\ path(Y,Z),\ not\ edge(X,Y),\ Z = P+1.}\\
\mathtt{:-\ path(X,n),\ path(Y,1),\ not\ edge(X,Y).}
\end{array}
}
\]
This program is usable for gringo with clasp, using the
\texttt{--shift} option (transforming the disjunctive rule into
several nondisjunctive ones), and DLV. We can observe that the
extensional definition is rewritten into a number of facts and that
the constant definition also just changes syntax. As for the
permutation statement, here we first use a predicate $\mathtt{d}$
representing the domain definition, and then a disjunctive rule and an
integrity constraint. The disjunctive rule states that each tuple in
the domain definition must be assigned one of the numbers 1 to 6, and
the integrity constraint enforces the bijection, that is, no different
tuples of the domain definition must be assigned the same number. The
final two integrity constraints are direct translations from the
NP-SPEC program. The only difference is the arithmetic expression that
has been moved outside the fact in order to conform to DLV's syntax
(gringo would also have accepted the immediate translation from the
NP-SPEC program).
\end{example}

\section{Translation from NP-SPEC to ASP}\label{sec:npspectoasp}

We now report how the various constructs of NP-SPEC programs can be translated into ASP. We start with the \DATABASE{} section constructs.
An extensional declaration of the form $\mathtt{p = \{ t_1, \ldots, t_n\}}$
will be translated to facts $\mathtt{p(t_1) \cdots p(t_n)}$,
and one of the form $\mathtt{p = \{ n..m \}}$ will be translated to
facts $\mathtt{p(n) \cdots p(m).}$
Constant declarations such as
$\mathtt{c = i}$, instead, will be managed in-memory.

Now for the main task, translating the \SPECIFICATION{} constructs. We first look at metafacts. The simplest one is $\mathtt{Subset(d,p).}$
The exact translation of this metafact (and all others as well) depends on how $\mathtt{d}$ is specified. First assume that it is an extensional predicate $\mathtt{d}$ of arity $\mathtt{n}$, then we can directly use it and produce
\[
\mathtt{p(X_1,\ldots,X_n) \lor -p(X_1,\ldots,X_n)\ :-\ d(X_1,\ldots,X_n).}
\]
Otherwise, we will create an extensional predicate $\mathtt{d}$ of appropriate arity (assuming that it is a fresh symbol) as follows:
\begin{itemize}
\item
for the Cartesian product $\mathtt{p >< q}$, the following set of facts is created:
$\{\mathtt{d(x_1,\ldots,}$ $\mathtt{x_{i+j}) \mid p(x_1,\ldots,x_i) \wedge q(x_{i+1},\ldots,x_{i+j})}\}$,
where $\mathtt{i}$ and $\mathtt{j}$ are the arities of $\mathtt{p}$ and $\mathtt{q}$, respectively;

\item
for the union $\mathtt{p + q}$, the following set of facts is created:
$\{\mathtt{d(x_1,\ldots,x_{i}) \mid p(x_1,\ldots,}$ $\mathtt{x_i) \vee q(x_{1},\ldots,x_{i})}\}$,
where $\mathtt{i}$ is the arity of both $\mathtt{p}$ and $\mathtt{q}$;

\item
for the intersection $\mathtt{p * q}$, the following set of facts is created:
$\{\mathtt{d(x_1,\ldots,x_{i}) \mid p(x_1,}$ $\mathtt{\ldots,x_i) \wedge q(x_{1},\ldots,x_{i})}\}$,
where $\mathtt{i}$ is the arity of both $\mathtt{p}$ and $\mathtt{q}$; and

\item
for the difference $\mathtt{p - q}$, the following set of facts is created:
$\{\mathtt{d(x_1,\ldots,x_{i}) \mid p(x_1,}$ $\mathtt{\ldots,x_i) \wedge \neg.q(x_{1},\ldots,x_{i})}\}$,
where $\mathtt{i}$ is the arity of both $\mathtt{p}$ and $\mathtt{q}$,
and $\mathtt{\neg.q(x_{1},\ldots,x_{i})}$ is true if and only if the fact
$\mathtt{q(x_{1},\ldots,x_{i})}$ is not part of the translation.
\end{itemize}
For nested domain definitions, we just repeat this process recursively using fresh symbols in each recursive step. In the following we will assume that domain definitions have been treated in this way and that the top-level predicate of the translation is $\mathtt{d}$.
If available (for instance when using gringo or lparse), we can also use choice rules for translating $\mathtt{Subset(d,p)}$:
\[
\begin{array}{l}
\mathtt{\{p(X_1,\ldots,X_n):d(X_1,\ldots,X_n)\}.}
\end{array}
\]

For the metafact $\mathtt{Permutation(d,p)}$, we will create
\[
\begin{array}{l}
\mathtt{p(X_1,\ldots, X_n,1) \vee \ldots \vee p(X_1,\ldots, X_n,c)\ :-\ d(X_1,\ldots, X_n).}\\
\mathtt{:-\  p(X_1,\ldots, X_n,A), p(Y_1,\ldots, Y_n,A), X_1 != Y_1. }\\
\qquad\vdots\\
\mathtt{:-\  p(X_1,\ldots, X_n,A), p(Y_1,\ldots, Y_n,A), X_n != Y_n. }\\
\end{array}
\]
where $\mathtt{n}$ is the arity of $\mathtt{d}$ and $\mathtt{c}$ is the cardinality of $\mathtt{d}$. The first rule specifies intuitively that for each tuple in $\mathtt{d}$ one of 
$\mathtt{p(X_1,\ldots,X_n,1)} \cdots \mathtt{p(X_1,\ldots,X_n,c)}$ should hold, and by minimality exactly one of these will hold. The integrity constraints ensure that no different numbers will be associated to the same tuple.
As an alternative to the disjunctive rule, one can use a choice rule
\[
\begin{array}{l}
\mathtt{1 \{p(X_1,\ldots,X_n,1..c)\} 1 :-\ d(X_1,\ldots,X_n).}
\end{array}
\]
Instead of the $\mathtt{n}$ integrity constraints it is possible to write just one using an aggregate, if available. In the DLV syntax, one could write
\[
\begin{array}{l}
\mathtt{:-\ \#count \{X_1,\ldots,X_n:p(X_1,\ldots,X_n,A)\}>1,\ p(\_,\ldots,\_,A).}\\
\end{array}
\]
or in gringo syntax
\[
\begin{array}{l}
\mathtt{:-\ 2\ \#count\{p(X1,\ldots,X_n,A)\},\ p(\_,\ldots,\_,A).}\\
\end{array}
\]

The remaining metafacts are actually much simpler to translate, as the bijection criterion does not have to be checked. For $\mathtt{Partition(d,p,k)}$, we will simply create
\[
\begin{array}{l}
\mathtt{p(X_1,\ldots, X_n,0) \vee \ldots \vee p(X_1,\ldots, X_n,k-1)\ :-\ d(X_1,\ldots, X_n).}\\
\end{array}
\]
or the respective choice rule
\[
\begin{array}{l}
\mathtt{1 \{p(X_1,\ldots,X_n,0..k-1)\} 1 :-\ d(X_1,\ldots,X_n).}
\end{array}
\]
where $\mathtt{n}$ is the arity of $\mathtt{d}$. For $\mathtt{IntFunc(d,p,i..j)}$,
we will simply create
\[
\begin{array}{l}
\mathtt{p(X_1,\ldots, X_n,i) \vee \ldots \vee p(X_1,\ldots, X_n,j)\ :-\ d(X_1,\ldots, X_n).}\\
\end{array}
\]
or the respective choice rule
\[
\begin{array}{l}
\mathtt{1 \{p(X_1,\ldots,X_n,i..j)\} 1 :-\ d(X_1,\ldots,X_n).}
\end{array}
\]
where $\mathtt{n}$ is the arity of $\mathtt{d}$.

What remains are the Datalog rules of the \SPECIFICATION{} section. Essentially, each
$\mathtt{Head\ \ruleimpl\ Body}$
is directly translated into
$\mathtt{Head'\ :-\ Body'}$,
with only minor differences. If $\mathtt{Head}$ is \fail{}, then
$\mathtt{Head'}$ is empty, otherwise it will be exactly the same. The
difference between $\mathtt{Body}$ and $\mathtt{Body'}$ is due to
different syntax for arithmetics, aggregates and due to safety
requirements. Concerning arithmetics, gringo can accept almost the
same syntax as NP-SPEC with only minor differences ($\mathtt{\#abs}$
instead of $\mathtt{abs}$, $\mathtt{\#pow}$ instead of $\texttt{\^{}}$),
while DLV is much more restrictive. DLV currently does not support
negative integers and it does not provide constructs corresponding to
$\mathtt{abs}$ and $\texttt{\^{}}$. Moreover, arithmetic expressions may
not be nested in DLV programs, but this limitation can be overcome by
flattening the expressions. 

Concerning aggregates, DLV and gringo
support similar syntax, which is a little bit different from the one
used in NP-SPEC but rather straightforward to rewrite
according to the following schema:
Arguments marked with asterisks are first replaced with fresh variables;
these are the arguments on which the aggregation function is applied.
Apart from $\mathtt{COUNT}$, exactly one asterisk may appear in each aggregate.
Hence, an aggregate $\mathtt{SUM(p(*,\_,Y),Z:n..m)}$
is written in DLV's syntax as
\[
\mathtt{\#sum\{X : p(X,\_,Y)\} = Z,\ d(Z)}
\]
and in gringo's syntax as
\[
\mathtt{Z\ \#sum[p(X,\_,Y) = X]\ Z,\ d(Z)}
\]
where $\mathtt{X}$ is a fresh variable and $\mathtt{d}$ is a fresh predicate defined by facts
$\mathtt{d(n) \cdots d(m).}$
Aggregates $\mathtt{MIN}$ and $\mathtt{MAX}$ are rewritten similarly, while an aggregate
\[
\mathtt{COUNT(p(*,\_,*,Y),Z:n..m)}
\]
is written in DLV's syntax as
\[
\mathtt{\#count\{X_1,X_2 : p(X_1,\_,X_2,Y)\} = Z,\ d(Z)}
\]
and in gringo's syntax by
\[
\mathtt{Z\ \#count\{p(X_1,\_,X_2,Y)\}\ Z,\ d(Z)}.
\]

A more difficult problem presents the
safety conditions enforced by the ASP systems. NP-SPEC has a fairly
lax safety criterion, while for instance DLV requires each variable to
occur in a positive, non-builtin body literal, and also gringo has a
similar criterion. This mismatch can be overcome by introducing
appropriate domain predicates when needed.

\section{Experiments}\label{sec:experiments}

We have created a prototype implementation of the transformation
described in section~\ref{sec:npspectoasp}, which is available at
\url{http://archives.alviano.com/npspec2asp/}. It is written in
\texttt{C++} using \texttt{bison} and \texttt{flex}, and called
NPSPEC2ASP. The implementation at the moment does only rudimentary
correctness checks of the program and is focussed on generating ASP
programs for correct NP-SPEC input. Moreover, at the moment it
generates only the disjunctive rules described in
section~\ref{sec:npspectoasp} rather than the choice rules, but we
plan to add the possibility to create variants of the ASP code in the
near future. For the experiments, the transformation used for 
$\mathtt{Permutation}$ produced the integrity constraint with the
counting aggregate.

We used this implementation to test the viability of our approach, in
particular assessing the efficiency of the proposed rewriting in ASP
with respect to the previously available transformation into SAT.  In
the benchmark we included several instances available on the NP-SPEC
site.  More specifically, we considered two sets of instances, namely
the \emph{miscellanea} and \emph{csplib2npspec} benchmarks.  Even if
these instances have been conceived for demonstrating the expressivity
of the language rather than for assessing the efficiency of an
evaluator, it turned out that even for these comparatively small
instances there are quite marked performance differences.
Below we provide some more details on the testcases in the \emph{miscellanea} benchmark.
\begin{itemize}
\item
\emph{Coloring} is an instance of the \emph{Graph Coloring} problem, i.e., given a graph $G$ and a set of $k$ colors, checking whether it is possible to assign a color to each node of $G$ in such a way that no adjacent nodes of $G$ share the same color.
The tested instance has 6 nodes and 3 colors.

\item
In the \emph{Diophantine} problem, three positive integers $a, b, c$ are given, and an integer solution to the equation $ax^2 + by = c$ is asked for.
The parameters of the tested instance are $a = 5$, $b = 3$, and $c = 710$.

\item
The \emph{Factoring} problem consists of finding two non-trivial factors (i.e., greater than $1$) of a given integer $n$.
In the tested instance, $n = 10000$.

\item
In the \emph{Hamiltonian Cycle} problem a graph $G$ is given, and a cycle traversing each node exactly once is searched.
The tested graph has 6 nodes.

\item
An instance of the \emph{Job Shop Scheduling} problem consists of integers $n$ (jobs), $m$ (tasks), $p$ (processors), and $D$ (global deadline).
Jobs are ordered collections of tasks, and each task is performed on a processor for some time.
Each processor can perform one task at a time, and the tasks belonging to the same job must be performed in order.  
The problem is checking whether it is possible for all jobs to meet deadline $D$.
In the testcase, $n=6$, $m=36$, $p=6$, and $D=55$.

\item
In the \emph{Protein Folding} problem, a sequence of $n$ elements in $\{H,P\}$ is given,
and the goal is to find a connected, non-overlapping shape of the sequence on a bi-dimensional, discrete grid, so that the number of ``contacts'', i.e., the number of non-sequential pairs of $H$ for which the Euclidean distance of the positions is 1, is in a given range $R$.
In the testcase, $n=6$, and $R = \{1..12\}$.

\item
In the \emph{Queens} problem, an integer $n$ is given, and the goal is to place $n$ non-attacking queens on a $n \times n$ chessboard.
In the tested instance, $n = 5$.

\item
Given an array $A$ of integers, the \emph{Sorting} problem consists of arranging the elements of $A$ in non-descending order.
In the tested instance, the array has 7 elements.

\item
An instance of the \emph{Subset Sum} problem comprises a finite set $A$, a size $s(a) \in \mathbb{N}^+$ for each $a \in A$, and $B \in \mathbb{N}^+$.
The goal of the problem is checking whether there is a subset $A'$ of $A$ such that the sum of the sizes of the elements in $A'$ is exactly $B$.
In the tested instance, set $A$ has 5 elements and $B = 10$.

\item
In a $Sudoku$, the goal is to fill a given (partially filled) grid with the numbers
1 to 9, so that every column, row, and $3\times3$ box indicated by slightly
heavier lines has the numbers 1 to 9.

\item
\emph{3-SAT} is a well-known NP-complete problem: Given a propositional formula $T$ in conjunctive normal form, in which each clause has exactly three literals, is $T$ satisfiable, i.e., does there exist an assignment of variables of $T$ to $\{\mathit{true},\mathit{false}\}$ that makes $T$ evaluate to $\mathit{true}$?
The tested instance has 3 clauses.

\item
The \emph{Tournament Scheduling} problem consists of assigning the matches to rounds of a round-robin tournament for a sports league.
The match is subject to several constraints, such as:
(i) complementary teams $t_1$ and $t_2$ have complementary schedules, i.e., for each round $r$,
if $t_1$ plays home in $r$ then $t_2$ plays away in $r$, and vice versa;
(ii) two top matches cannot take place at distance smaller than a given value;
(iii) any team cannot match two top teams at distance smaller than a given value.
(See \cite{cado-scha-2005-aij} for details.)
The tested instance has 6 teams.
\end{itemize}
Below we describe the testcases in the \emph{csplib2npspec} benchmark.
\begin{itemize}
\item
Given $n \in \mathbb{N}$, find a vector $s = (s_1, ..., s_n)$ such that
(i) $s$ is a permutation of $Z_n = \{0,1,\ldots,n-1\}$; and 
(ii) the interval vector $v = (|s_2-s_1|, |s_3-s_2|, \ldots, |s_n-s_{n-1}|)$ is a permutation of $Z_n\setminus\{0\} = \{1,2,\ldots,n-1\}$.
A vector $v$ satisfying these conditions is called an all-interval series of size $n$; the problem of finding such a series is the \emph{All-interval Series} problem of size $n$.
In the tested instance, $n = 20$.

\item
In the \emph{BACP} (balanced academic curriculum problem), each course has associated a number of credits and can have other courses as prerequisites.
The goal is to assign a period to every course in a way that the number of courses and the amount of credits per period are in given ranges, and the prerequisite relationships are satisfied.
The tested instance comprises 7 courses and 2 periods.

\item
A \emph{BIBD} is defined as an arrangement of $v$ distinct objects into $b$ blocks such that each block contains exactly $k$ distinct objects, each object occurs in exactly $r$ different blocks, and every two distinct objects occur together in exactly $\lambda$ blocks.
The parameters of the tested instance are $v = 7$, $b = 7$, $k = 3$, $r = 3$, and $\lambda = 1$.

\item
In the \emph{Car Sequencing} problem, a number of cars are to be produced; they are not identical, because different
options are available as variants on the basic model.  
The assembly line has different stations which install the various options
(air-conditioning, sun-roof, etc.). These stations have been designed to handle
at most a certain percentage of the cars passing along the assembly
line. Consequently, the
cars must be arranged in a sequence so that the capacity of each station is
never exceeded.
In the testcase there are 10 cars, 6 variants on a basic model, and 5 options.

\item
A \emph{Golomb ruler} is a set of $m$ integers $0 = a_1 < a_2 < \cdots < a_m$ such that the $m(m-1)/2$
differences $a_j - a_i$ ($1 \leq i < j \leq m)$ are distinct.
In the tested instance, $m = 8$ and $a_m$ must be lesser than or equals to 10.

\item
\emph{Langford}'s problem is to arrange $k$ sets of
numbers 1 to $n$ so that each appearance of the number $m$ is $m$ numbers on from
the last.
In the tested instance, $k = 3$ and $n = 9$.

\item
Given integers $n$ and $b$, the objective of the \emph{Low Autocorrelation} problem is to construct a binary sequence $S_i$ of length $n$, where each bit takes the value +1 or -1, so that $E = \sum_{k=1}^{n-1} (C_k)^2 \leq b$, where $C_k = \sum_{i=0}^{n-k-1} S_i \cdot S_{i+k}$.
In the tested instance, $n = 5$ and $b = 2$.

\item
An order $n$ \emph{magic square} is a $n \times n$ matrix containing the numbers 1 to $n^2$,
with each row, column and main diagonal summing up to the same value.
In our setting, $n=3$.

\item
The \emph{Ramsey} problem is to color the edges of a complete graph with $n$ nodes using at most $k$ colors, in such a way that there is no monochromatic triangle in the graph.
In the tested instance, $n=5$ and $k=3$.

\item
The \emph{Round-robin Tournament} problem is to schedule a tournament of $n$ teams over $n-1$ weeks, with each week divided into $n/2$ periods, and each period divided into two slots.
A tournament must satisfy the following three constraints:
every team plays once a week;
every team plays at most twice in the same period over the tournament;
every team plays every other team.
In our setting, $n=4$.

\item
\emph{Schur's Lemma} problem is to put $n$ balls labelled $\{1,\ldots,n\}$ into 3 boxes so that for any triple of balls $(x,y,z)$ with $x+y=z$, not all are in the same box.
In the tested instance, $n=10$.

\item
In the \emph{Social Golfer} problem there are $n$ golfers, each of whom play golf once a week, and always in groups of $s$.
The goal is to determine a schedule of play for these golfers, to last $l$ weeks, such
that no golfer plays in the same group as any other golfer on more than one occasion.
In our setting, $n=8$, $s=2$, and $l=4$.
\end{itemize}

The experiment has been executed on an Intel Core2 Duo P8600 2.4 GHz with 4 GB of central memory, running Linux Mint Debian Edition (wheezy/sid) with kernel Linux 3.2.0-2-amd64.
The tools SPEC2SAT and NPSPEC2ASP have been compiled with gcc 4.6.3.
The other tools involved in the experiment are
satz 215.2 \cite{li-99-satz},
minisat 1.14 \cite{sore-nikl-2003},
gringo 3.0.4 \cite{gebs-etal-2007-lpnmr},
clasp 2.0.6 \cite{gebs-etal-2007-ijcai}, and
DLV 2011-12-21 \cite{leon-etal-2002}.

In our experiment, we first measured the running time required by SPEC2SAT and NPSPEC2ASP to rewrite the input specification into SAT and ASP, respectively.
Then, for each SAT encoding produced by SPEC2SAT, we ran three SAT solvers, namely satz, minisat and clasp, to obtain one solution if one exists.
For each of these executions we measured the time to obtain the solution or the assertion that none exists, thus the sum of the running times of SPEC2SAT and of the SAT solvers.
Moreover, for each ASP encoding produced by NPSPEC2ASP, we ran two instantiators, namely gringo and DLV (with option \verb!--instantiate!).
For each of these runs we measured the time required to compute the ground ASP program, thus the sum of the running times of NPSPEC2ASP and of the instantiator.
Finally, for each ground ASP program computed by gringo and DLV, we computed one solution by using clasp and DLV, respectively, and measured the overall time required by the tool-chain. We have also measured the sizes of the instantiated formulas and programs. For SPEC2SAT, we report the number of clauses in the produced formula and the number of propositional variables occurring in it. For DLV and gringo we report the number of ground rules produced and the number of ground atoms occurring in them. There is a slight difference in the statistics provided by DLV and gringo: DLV does not count ground atoms (and facts) that were already found to be true; to be more comparable, we added the number of facts for DLV.

Experimental results concerning the \emph{miscellanea} benchmark are reported in Table~\ref{tab:experiment}.
We first observe that the time required by NPSPEC2ASP is below the measurement accuracy, while the execution time of SPEC2SAT is higher, sometimes by several orders of magnitude.
In fact, SPEC2SAT has to compute a ground SAT instance to pass to a SAT solver, while NPSPEC2ASP outputs a non-ground ASP program.
A fairer comparison is obtained by adding to the time taken by NPSPEC2ASP the time required by the ASP instantiator to obtain a ground ASP program.
Columns gringo and ``DLV inst'' report these times, which are however always less than those of SPEC2SAT.
In Table~\ref{tab:experiment_size} it can be seen that also the number of ground rules produced by the ASP systems is usually smaller than the number of clauses produced by SPEC2SAT, even if often the number of ground atoms exceeds the number of propositional variables. 

\begin{table}[b!]
 \figrule
 \caption{Running times on the \emph{miscellanea} benchmark}\label{tab:experiment}
 \vspace{-2em}
\begin{center}
\begin{tabular}{|c|rrrr|rrrrr|}
\hline
\multirow{3}{*}{Instance} & \multicolumn{4}{c|}{SPEC2SAT} & \multicolumn{5}{c|}{NPSPEC2ASP}\\
\cline{2-10}
 & \multirow{2}{*}{only} & \multirow{2}{*}{satz} & \multirow{2}{*}{minisat} & \multirow{2}{*}{clasp} & \multirow{2}{*}{only} & DLV & \multirow{2}{*}{DLV} & \multirow{2}{*}{gringo} & gringo \\
 & & & & & & inst & & & \verb!+!clasp \\
\hline
coloring & 0.01 & 0.01 & 0.06 & 0.01 & 0.00 & 0.00 & 0.00 & 0.00 & 0.01\\
diophantine & 0.75 & 0.81 & 0.77 & 0.79 & 0.00 & 0.05 & 0.04 & 0.04 & 0.06\\
factoring & 5.99 & 10.21 & 6.09 & 7.15 & 0.00 & 0.23 & 0.38 & 0.17 & 1.05\\
hamiltonianCycle & 0.03 & 0.03 & 0.03 & 0.03 & 0.00 & 0.01 & 0.01 & 0.01 & 0.01\\
jobShopScheduling & 43.39 & 44.89 & 44.64 & 44.34 & 0.00 & 1.63 & 87.96 & 1.01 & 1.92\\
proteinFolding & 132.17 & \verb!>!600 & 165.03 & 134.49 & 0.00 & N/A$^{*}$ & N/A$^{*}$ & 2.51 & 4.18\\
queens & 0.03 & 0.04 & 0.03 & 0.04 & 0.00 & 0.01 & 0.01 & 0.01 & 0.01\\
sorting & 0.03 & 0.03 & 0.03 & 0.03 & 0.00 & 0.01 & 0.01 & 0.01 & 0.01\\
subsetSum & 0.11 & 0.11 & 0.10 & 0.11 & 0.00 & 0.00 & 0.00 & 0.00 & 0.01\\
sudoku & 3.05 & 3.07 & 3.06 & 3.04 & 0.00 & 0.11 & 0.11 & 0.11 & 0.17\\
threeSat & 0.01 & 0.01 & 0.01 & 0.01 & 0.00 & N/A$^{*}$ & N/A$^{*}$ & 0.01 & 0.01\\
tournamentScheduling & 0.45 & 0.45 & 0.45 & 0.45 & 0.00 & 0.04 & 0.03 & 0.02 & 0.02\\
\hline
\end{tabular}
\end{center}
\ \vspace{-1em}\\
$\phantom{xxxx}^{*\phantom{*}}$ The instance contains negative integers.

 \figrule
\end{table}

\begin{table}[b!]
 \figrule
 \caption{Instance sizes of the \emph{miscellanea} benchmark}\label{tab:experiment_size}
 \vspace{-2em}
\begin{center}
\begin{tabular}{|c|rr|rr|rr|}
\hline
\multirow{3}{*}{Instance} & \multicolumn{2}{c|}{\multirow{2}{*}{SPEC2SAT}} & \multicolumn{4}{c|}{NPSPEC2ASP}\\
\cline{4-7}
 & & & \multicolumn{2}{c|}{DLV} & \multicolumn{2}{c|}{gringo}\\
\cline{2-7}
 & Clauses & Variables & Rules & Atoms & Rules & Atoms\\
\hline
coloring & 45 & 18 & 40 & 31 & 58 & 38\\
diophantine & 14,628 & 140 & 9,800 & 142 & 9,940 & 145\\
factoring & 123,748 & 498 & 61,998 & 500 & 62,496 & 503\\
hamiltonianCycle & 348 & 36 & 261 & 99 & 291 & 94\\
jobShopScheduling & 209,495 & 1,980 & 156,107 & 2,052 & 158,087 & 2,089\\
proteinFolding & 735,721 & 669 & N/A$^{*}$ & N/A$^{*}$ & 520,107 & 347\\
queens & 165 & 25 & 125 & 65 & 145 & 61\\
sorting & 427 & 49 & 252 & 126 & 294 & 120\\
subsetSum & 1,418 & 125 & 49 & 54 & 100 & 77\\
sudoku & 33,825 & 1,458 & 24,777 & 2,545 & 25,263 & 1,736\\
threeSat & 30 & 39 & N/A$^{*}$ & N/A$^{*}$ & 87 & 76\\
tournamentScheduling & 1,641 & 108 & 1,675 & 115 & 1,810 & 182\\
\hline
\end{tabular}
\end{center}
\ \vspace{-1em}\\
$\phantom{xxxx}^{*\phantom{*}}$ The instance contains negative integers.

 \figrule
\end{table}

Concerning the computation of one solution from each ground specification, all considered SAT and ASP solvers are fast in almost all tests.
The only exceptions are satz for \emph{proteinFolding}, which exceeds the allotted time, and DLV for \emph{jobShopScheduling}, whose execution lasted around 88 seconds.
We also note that DLV has not been tested on 2 instances containing negative integers, which are not allowed in the DLV language.

Table~\ref{tab:experiment2} reports experimental results concerning the \emph{csplib2npspec} benchmark.
We start by observing that instances in this benchmark are more resource demanding than instances in the \emph{miscellanea} benchmark.
In fact, we note that \emph{golombRuler} is too difficult for SPEC2SAT, which did not terminate on the allotted time on this instance.
On the other hand, the rewriting provided by NPSPEC2ASP is processed in around 28 seconds by gringo+clasp and in around 24 seconds by DLV.
Another hard instance is \emph{allInterval}, for which only satz and DLV terminated in the allotted time. 
All other solvers, including gringo+clasp, exceeded the allotted time, even if the NPSPEC2ASP rewriting and the instantiation by gringo is produced in less time than the output of SPEC2SAT. This instance is an outlier in our experiments. 
In almost all other instances the ASP solvers compute solutions in less than 1 second, while SAT solvers typically require several seconds, see in particular \emph{langford}, \emph{magicSquare} and \emph{lowAutocorrelation}.
 The size of the programs produced by the ASP instantiators is always smaller than the size of the formulas produced by SPEC2SAT, sometimes by orders of magnitude, even if the number of ground atoms often exceeds the number of propositional variables. A major cause for the difference in size appear to be aggregates in the problem specification, which are supported natively by ASP systems, but require expensive rewritings for SAT solvers.

\begin{table}[b!]
 \figrule
 \caption{Running times on the \emph{csplib2npspec} benchmark}\label{tab:experiment2}
 \vspace{-2em}
\begin{center}
\begin{tabular}{|c|rrrr|rrrrr|}
\hline
\multirow{3}{*}{Instance} & \multicolumn{4}{c|}{SPEC2SAT} & \multicolumn{5}{c|}{NPSPEC2ASP}\\
\cline{2-10}
 & \multirow{2}{*}{only} & \multirow{2}{*}{satz} & \multirow{2}{*}{minisat} & \multirow{2}{*}{clasp} & \multirow{2}{*}{only} & DLV & \multirow{2}{*}{DLV} & \multirow{2}{*}{gringo} & gringo \\
 & & & & & & inst & & & \verb!+!clasp \\
\hline
allInterval & 1.43 & 36.98 & \verb!>!600 & \verb!>!600 & 0.00 & 0.07 & 0.88 & 0.06 & \verb!>!600\\
bacp & 6.15 & 6.33 & 6.18 & 6.22 & 0.00 & 0.01 & 0.01 & 0.01 & 0.01\\
bibd & 3.98 & 4.20 & 4.03 & 4.06 & 0.00 & 0.03 & 0.11 & 0.03 & 0.03\\
carSequencing & 8.69 & 14.26 & 8.82 & 8.86 & 0.00 & 0.87 & 0.83 & 0.34 & 0.48\\
golombRuler & \verb!>!600 & \verb!>!600 & \verb!>!600 & \verb!>!600 & 0.00 & 23.20 & 23.70 & 26.93 & 28.19\\
langford & 11.57 & 12.58 & 12.28 & 12.62 & 0.00 & 0.04 & 0.90 & 0.03 & 0.09\\
lowAutocorrelation & 23.17 & 24.02 & 23.15 & 23.36 & 0.00 & N/A$^{*}$ & N/A$^{*}$ & 0.03 & 0.03\\
magicSquare & 10.55 & 10.74 & 10.59 & 10.54 & 0.00 & 0.17 & 21.94 & 0.12 & 0.32\\
ramseyProblem & 0.02 & 0.02 & 0.02 & 0.02 & 0.00 & 0.00 & 0.00 & 0.01 & 0.01\\
roundrobinTournament & 2.11 & 2.27 & 2.15 & 2.14 & 0.00 & 0.02 & 0.02 & 0.01 & 0.02\\
schursLemma & 0.13 & 0.14 & 0.14 & 0.14 & 0.00 & 0.01 & 0.01 & 0.01 & 0.01\\
socialGolfer & 7.32 & 7.45 & 7.52 & 7.52 & 0.00 & 0.09 & 0.14 & 0.05 & 0.07\\
\hline
\end{tabular}
\end{center}
\ \vspace{-1em}\\
$\phantom{xxxx}^{*\phantom{*}}$ The instance contains negative integers.

 \figrule
\end{table}

\begin{table}[b!]
 \figrule
 \caption{Instance sizes of the \emph{csplib2npspec} benchmark}\label{tab:experiment_size2}
 \vspace{-2em}
\begin{center}
\begin{tabular}{|c|rr|rr|rr|}
\hline
\multirow{3}{*}{Instance} & \multicolumn{2}{c|}{\multirow{2}{*}{SPEC2SAT}} & \multicolumn{4}{c|}{NPSPEC2ASP}\\
\cline{4-7}
 & & & \multicolumn{2}{c|}{DLV} & \multicolumn{2}{c|}{gringo}\\
\cline{2-7}
 & Clauses & Variables & Rules & Atoms & Rules & Atoms\\
\hline
allInterval & 21,737 & 761 & 9,239 & 1,639 & 9,961 & 1,601\\
bacp & 39,531 & 1,518 & 314 & 316 & 436 & 360\\
bibd & 31,843 & 4,424 & 2,684 & 2,047 & 4,091 & 2,279\\
carSequencing & 39,875 & 786 & 33,398 & 219 & 33,506 & 218\\
golombRuler & N/A$^{**}$ & N/A$^{**}$ & 653,593 & 96 & 1,149,561 & 105\\
langford & 130,518 & 7299 & 3,736 & 793 & 4,015 & 803\\
lowAutocorrelation & 186,407 & 5,952 & N/A$^{*}$ & N/A$^{*}$ & 2,339 & 1,041\\
magicSquare & 38,564 & 1,975 & 5458 & 872 & 18,445 & 14,513\\
ramseyProblem & 80 & 30 & 60 & 50 & 90 & 61\\
roundrobinTournament & 9,272 & 456 & 1,203 & 275 & 1,467 & 400\\
schursLemma & 175 & 30 & 155 & 40 & 185 & 51\\
socialGolfer & 21,600 & 1,424 & 11,097 & 441 & 11,321 & 442\\
\hline
\end{tabular}
\end{center}
\ \vspace{-1em}\\
$\phantom{xxxx}^{*\phantom{*}}$ The instance contains negative integers.\\
$\phantom{xxxx}^{**}$ The system did not terminate in 30 minutes.

 \figrule
\end{table}

The experimental results show that translating NP-SPEC programs into
ASP rather than SAT seems to be preferable, due to the fact that
sophisticated instantiation techniques can be leveraged. Moreover,
also the nondeterministic search components of ASP systems can compete
well with SAT solvers, making the use of ASP solvers very attractive
for practical purposes.

\section{Conclusion}\label{sec:concl}
In this paper we have presented a transformation of NP-SPEC programs
into ASP. The translation is modular and not complex at all, allowing
for very efficient transformations. Compared to the previously
available transformation into Boolean satisfiability, there are a
number of crucial differences: While our transformation is from a
formalism with variables into another formalism with variables,
Boolean satisfiability of course does not allow for object
variables. Therefore any transformation to that language has to do an
implicit instantiation. It is obvious that instantiation can be very
costly, and thus using sophisticated instantiation methods is often
crucial. However, optimization methods for instantiation are often
quite involved and not easy to implement, and therefore adopting them
in a transformation is detrimental. After all, the appeal of
transformations are usually their simplicity and the possibility to
re-use existing software after the transformation. Our transformation
method does just that; by not instantiating it is possible to re-use
existing instantiators inside ASP systems, many of which use quite
sophisticated techniques like join ordering heuristics, dynamic
indexing and many more. We have provided a prototype implementation
that showcases this advantage. Even if only rather small examples were
tested, already in most of those cases a considerable advantage of our
method can be observed. 

There is a second aspect of our work, which regards ASP. As can be
seen in section~\ref{sec:npspectoasp}, the translation of
$\mathtt{Permutation}$ either gives rise to possibly many integrity
constraints or one with an aggregate. In any case, all current ASP
instantiators will materialize all associations between tuples of the
domain definition and the permutation identifiers, even if the
identifiers are not really important for solving the problem. This
means that there are obvious symmetries in the instantiated
program. There exist proposals for symmetry breaking in ASP
(e.g.\ \cite{dres-etal-2011-aicom}), but they typically employ
automorphism detection. We argue that in cases like this, a statement
like $\mathtt{Permutation}$, $\mathtt{Partition}$, or
$\mathtt{IntFunc}$ would make sense as a language addition for ASP
solvers, which could exploit the fact that the permutation identifiers
introduce a particular known symmetry pattern that does not have to be
detected by any external tool.

Future work consists of consolidating the prototype software and
extending the experimentation. Moreover, we intend to investigate the
possibility to extend our transformation to work with other languages
that are similar to NP-SPEC. Finally, we also intend to explore the
possibility and impact of introducing $\mathtt{Permutation}$,
$\mathtt{Partition}$, or $\mathtt{IntFunc}$ into ASP languages.

\bibliographystyle{splncs03}
\bibliography{aspocp2012}

\end{document}